\documentclass[conference]{IEEEtran}
\IEEEoverridecommandlockouts

\usepackage{cite}
\usepackage{amsmath,amssymb,amsfonts}
\usepackage{graphicx}
\usepackage{textcomp}
\usepackage{xcolor}
\usepackage{algorithm}
\usepackage{algpseudocode}

\usepackage{nicefrac} 
\newcommand{\labeledNodeSet}{\mathcal{V}^{L}}

\newcommand{\unLabeledNodeSet}{\mathcal{V}^{U}}

\usepackage{multirow}
\usepackage{float}
\usepackage{stfloats}
\usepackage{microtype}
\usepackage{subfigure}
\usepackage{booktabs} 
\usepackage{colortbl}
\usepackage{arydshln} 

\def\BibTeX{{\rm B\kern-.05em{\sc i\kern-.025em b}\kern-.08em
    T\kern-.1667em\lower.7ex\hbox{E}\kern-.125emX}}
\begin{document}

\title{ProGMLP: A Progressive Framework for GNN-to-MLP Knowledge Distillation with Efficient Trade-offs}


\author{Weigang~Lu$^{1,2}$,
		Ziyu~Guan$^{1}$,
		Wei~Zhao$^1$,
		Yaming~Yang$^1$,
		Yujie~Sun$^1$,
		Zheng~Liang$^2$,
        Yibing~Zhan$^3$,
        Dapeng~Tao$^4$\\ 
        \textit{$^1$School of Computer Science and Technology, Xidian University, Xi'an, China}\\ 
        \textit{$^2$ Department of Civil and Environmental Engineering, The Hong Kong University of Science and Technology,}\\
        \textit{Hong Kong SAR}\\
        \textit{$^3$JD Explore Academy, Beijing, China}\\ 
        \textit{$^4$School of Information Science and Engineering, Yunnan University, Kunming, China}\\ 
        \{wglu@stu., zyguan@, ywzhao@mail., yym@, yujsun@stu.\}xidian.edu.cn, \\
        liangz@ust.hk, zhanyibing@jd.com, dapeng.tao@gmail.com    
        }
        
%
\maketitle

\begin{abstract}
GNN-to-MLP (G2M) methods have emerged as a promising approach to accelerate Graph Neural Networks (GNNs) by distilling their knowledge into simpler Multi-Layer Perceptrons (MLPs). These methods bridge the gap between the expressive power of GNNs and the computational efficiency of MLPs, making them well-suited for resource-constrained environments. However, existing G2M methods are limited by their inability to flexibly adjust inference cost and accuracy dynamically, a critical requirement for real-world applications where computational resources and time constraints can vary significantly. To address this, we introduce a \underline{Pro}gressive framework designed to offer flexible and on-demand trade-offs between inference cost and accuracy for \underline{G}NN-to-\underline{MLP} knowledge distillation (ProGMLP). ProGMLP employs a Progressive Training Structure (PTS), where multiple MLP students are trained in sequence, each building on the previous one. Furthermore, ProGMLP incorporates Progressive Knowledge Distillation (PKD) to iteratively refine the distillation process from GNNs to MLPs, and Progressive Mixup Augmentation (PMA) to enhance generalization by progressively generating harder mixed samples. Our approach is validated through comprehensive experiments on eight real-world graph datasets, demonstrating that ProGMLP maintains high accuracy while dynamically adapting to varying runtime scenarios, making it highly effective for deployment in diverse application settings.
\end{abstract}


\section{Introduction}
GNN-to-MLP (\textbf{G2M}) methods have recently gained attention as an effective approach for accelerating the inference of Graph Neural Networks (GNNs)~\cite{gcn,gat,sage,sgc,gin,appnp,yang2022graph,skipnode} by distilling their knowledge into simpler Multi-Layer Perceptrons (MLPs). These methods typically involve training a single MLPs~\cite{glnn,nosmog,krd,ffg2m}, or a set of MLPs~\cite{adagmlp}, to mimic GNN predictions. They aim to bridge the gap between the expressive power of GNNs and the computational efficiency of MLPs, making them promising for deployment in resource-constrained environments or latency-sensitive applications. By pre-compiling graph knowledge into MLPs, G2M techniques significantly reduce the need for explicit graph traversal and neighborhood aggregation during inference, leading to faster prediction times.

\begin{figure}
	\includegraphics[width=1\columnwidth]{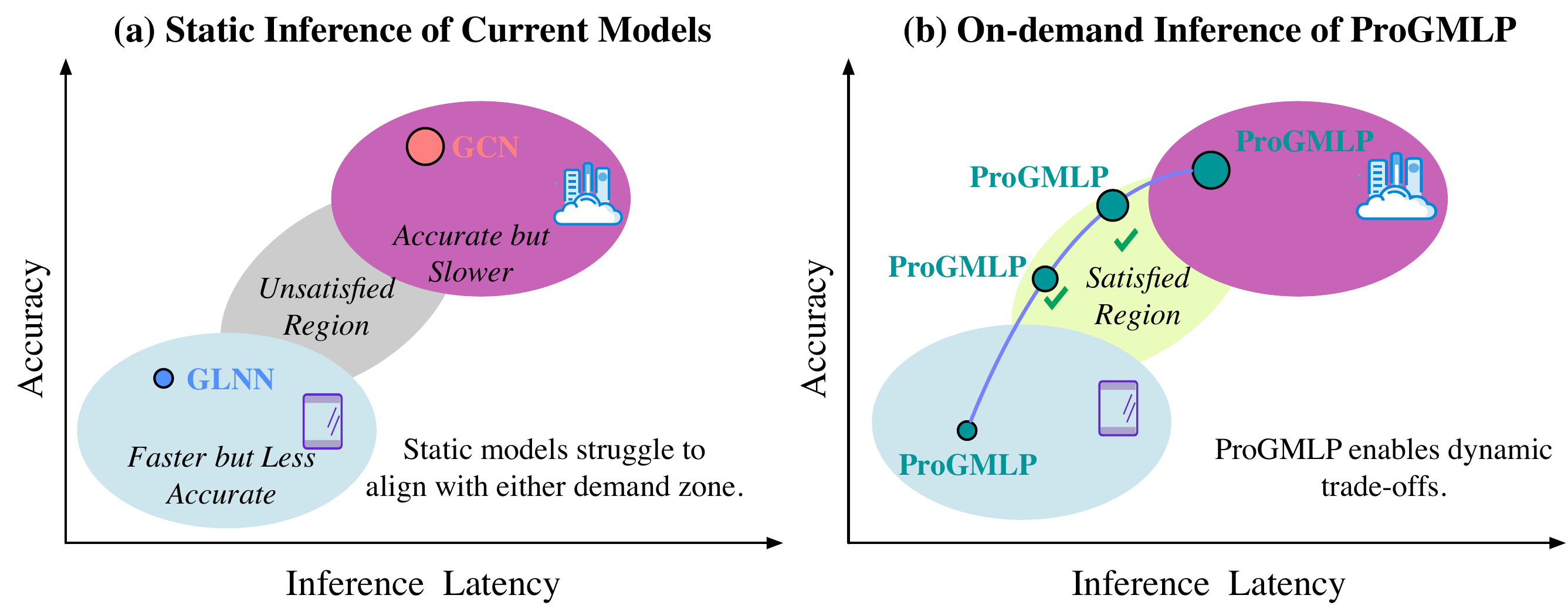}
	\caption{Illustration of the core motivation for ProGMLP. (a) Existing methods result in static models (e.g., a fast but less accurate GLNN, or a slow but accurate GNN) that are deployed for a single performance point. This creates a fundamental mismatch with the dynamic nature of real-world application demands, leaving a significant range of operational requirements unsatisfied. (b) In contrast, ProGMLP is deployed as a single, adaptive framework. It enables on-demand trade-offs by providing a spectrum of operating points along the accuracy-latency curve, allowing it to satisfy diverse requirements such as low-latency inference for edge devices or high-accuracy predictions on servers. \label{fig:intro}}
\end{figure}

\emph{However, a critical limitation of existing G2M approaches is their lack of flexibility in balancing inference cost and accuracy.} Current methods are designed to operate within fixed computational budgets, which means they cannot dynamically adjust to the varying demands of different application scenarios. As illustrated in Figure~\ref{fig:intro} (a), this creates a fundamental mismatch. System operators are forced to make a pre-emptive choice: deploy a fast, low-accuracy model suitable for edge devices, or a slow, high-accuracy model for server-side analytics. In real-world settings, the ability to control the trade-off between inference accuracy and computational cost on-demand during execution is crucial. For instance, in edge computing~\cite{edge_computing1,edge_computing2,edge_computing3,edge_computing4} or mobile environments~\cite{mobile1,mobile2}, where computational power and energy resources can fluctuate, a model that can adaptively tune its inference process to optimize for either speed or accuracy would be highly beneficial.  During peak loads, it may be necessary to serve predictions with lower latency at the cost of a slight accuracy drop to meet throughput demands. 	This reality necessitates the use of early-exit~\cite{early_exit1,early_exit2} or anytime inference~\cite{anytime1,anytime2,anytime3} applications, where the inference process may be interrupted due to changing resource availability. Unfortunately, this aspect of flexibility has been largely overlooked in existing G2M methods, limiting their practical utility.

To address this gap, we propose \textbf{ProGMLP}, a progressive framework designed to offer flexible and on-demand trade-offs between inference cost and accuracy in the context of G2M. The motivation behind ProGMLP is to enable users to dynamically adjust the inference process based on the specific needs of their applications, whether it be maximizing accuracy under loose time constraints or minimizing inference time when computational resources are scarce. As visualized in Figure~\ref{fig:intro} (b), ProGMLP is not a single-point solution but an adaptive framework that spans a spectrum of operating points on the accuracy-latency curve. ProGMLP achieves this through a sequence of progressively trained MLPs, each building on the knowledge of the previous one, and incorporates a performance-time budgeting mechanism that intelligently manages when to stop inference based on real-time performance metrics.

Our contributions can be summarized as follows:

\begin{itemize}
	\item \textbf{Pioneering Flexible Trade-offs in G2M Context:} We introduce a novel framework, ProGMLP, which is the first to explicitly address the need for flexible trade-offs between accuracy and inference cost within the G2M paradigm. While traditional G2M methods focus on static inference models, ProGMLP allows for dynamic adjustments at runtime, enabling users to optimize performance based on the specific demands of their application, such as varying computational resources or time constraints. This flexibility is particularly crucial for on-device inference, where the balance between speed and accuracy can significantly impact the usability and efficiency of the deployed model.
	\item \textbf{Novel Design with Progressive Enhancements:} ProGMLP incorporates several novel designs in its training process, including a Progressive Training Structure, Progressive Knowledge Distillation, and Progressive Mixup Augmentation, ensuring that each MLP student model is progressively refined and better suited for diverse tasks. 
	\item \textbf{Comprehensive Evaluation on Real-World Graphs:} We validate the effectiveness of ProGMLP through rigorous experiments on eight real-world graph datasets with different scales. Our evaluation demonstrates that ProGMLP not only maintains high accuracy but also provides the flexibility to adjust inference costs according to different runtime scenarios. This extensive experimentation highlights the practical utility of our approach and confirms that ProGMLP can be effectively deployed on real-word scenarios, offering both efficiency and performance.
\end{itemize}

\section{Backgrounds}

\subsection{Graph Neural Networks}
Graph Neural Networks (GNNs)~\cite{gcn,gat,sage,sgc,gin,appnp,yang2022graph,gcnii,jknet} are a class of models that leverage the graph structure to learn node representations. A GNN typically follows an iterative message-passing paradigm, where each node in the graph aggregates information from its neighboring nodes to update its own representation. Through multiple layers, GNNs enable the learning of node embeddings that incorporate not only the node’s features but also the structural information from its local and extended neighborhood. This mechanism allows GNNs to generalize well to diverse tasks such as node classification~\cite{skipnode,pcl,nodemixup}, link prediction~\cite{lpformer}, and graph classification~\cite{san}. However, despite their efficacy, GNNs face several challenges, particularly in terms of scalability and inference efficiency. As the size and complexity of real-world graphs grow, GNNs become increasingly expensive to train and deploy due to the high computational and memory costs associated with processing large graphs and multiple layers of message passing. This issue is further exacerbated when GNNs are deployed in resource-constrained environments, such as edge devices or mobile platforms, where both computation and memory are limited.

Most GNNs operate under a message-passing scheme, where each node's representation $h_{i}$ is iteratively updated by aggregating information from its neighbors. In the $l$-th layer, the representation $h^{(l)}_{i}$ is computed by first aggregating messages from neighboring nodes, followed by an update operation. This process can be formally expressed as:
\begin{align}
		&\tilde{h}^{(l)}_{i} = \operatorname{AGGREGATE}^{(l)}(\{h^{(l-1)}_{i} : i \in \mathcal{N}_{i}\}) \\
		&h^{(l)}_{i} = \operatorname{UPDATE}^{(l)}(\tilde{h}^{(l)}_{i}, h^{(l-1)}_{i}),
\end{align}
where $\mathcal{N}_{i}$ denotes the set of neighbors of node $i$, and $\operatorname{AGGREGATE}$ and $\operatorname{UPDATE}$ are the aggregation and update functions, respectively.

\subsection{GNN-to-GNN Knowledge Distillation} 
To address the efficacy issue in GNNs, GNN-to-GNN Knowledge Distillation (G2G KD)~\cite{lassance2020deep,zhang2023iterative,ren2021multi,joshi2022representation,wu2022knowledge,zhang2019your,ren2021multi,gnn-sd} has been extensively studied. They aim to compress large, complex Graph Neural Networks (GNNs) into smaller, more efficient GNN models. These approaches leverage KD techniques~\cite{kd_kl, kd_l2} to transfer the knowledge embedded in a large teacher GNN to a smaller student GNN. For example, methods like LSP~\cite{lsp} and TinyGNN~\cite{tinygnn} facilitate the transfer of localized structural information from teacher GNNs to their student counterparts. Similarly, RDD~\cite{rdd} enhances G2G KD by considering the reliability of nodes and edges. \emph{Although effective, these methods often require neighbor fetching during inference}, which can introduce latency and make them less practical for real-time or resource-constrained applications.

\subsection{GNN-to-MLP Knowledge Distillation} 
To address the inference efficacy issue in G2G KD, GNN-to-MLP (G2M) KD has emerged as a promising alternative. This approach transfers the knowledge of a GNN to a simpler MLPs model, which eliminates the need for message passing during inference and thus significantly reduces latency. Early work such as GLNN~\cite{glnn} introduces a general G2M framework where an MLPs student is trained using both ground-truth and soft labels from a GNN teacher. Subsequent methods like KRD~\cite{krd} introduces a reliable sampling strategy to improve the quality of the knowledge transferred, while NOSMOG~\cite{nosmog} and GSDN~\cite{gsdn} incorporates structural  information to further enhance the MLP student's performance. VQGraph~\cite{vqgraph} proposes to learn a codebook that represents informative local structures, and uses these local structures as additional information for distillation.  The most related work is AdaGMLP~\cite{adagmlp} which adopts an AdaBoosting ensemble framework to train multiple MLPs, collecting all the knowledge from each student to make predictions. 

However, while G2M methods have made notable progress in improving inference efficiency, they generally overlook a critical aspect: \emph{flexible inference}. They are designed for static computation budgets, where the trade-off between inference cost and accuracy is fixed after training. This lack of flexibility means these models cannot dynamically adapt to varying computational resources or application requirements in real-time, limiting their effectiveness in scenarios like edge computing or mobile environments where such adaptability is crucial. This rigidity poses a significant limitation, as it prevents G2M methods from optimizing performance on-the-fly, hindering their practical utility in diverse and changing environments.

\section{Methodology}
\subsection{Motivation}
In resource-constrained environments, it is often crucial to balance the trade-off between computational cost and model accuracy. For example, early exit strategies~\cite{early_exit1,early_exit2,early_exit3,early_exit4} have been extensively studied to allow models to terminate inference early if a satisfactory prediction can be made to improve energy efficiency. This is particularly useful in scenarios where the computational budget can be variable, but still allows the model to make high-confidence predictions in cases where the input is easy to classify. Early exit mechanisms are proactive in reducing computation by identifying simpler cases early in the inference process, improving efficiency without significantly sacrificing accuracy.

On the other hand, anytime inference strategies~\cite{anytime1,anytime2,anytime3} have also been proposed to provide flexible, interruptible inference, where a model can yield a prediction even when inference is interrupted, e.g., due to the time limitation. In this scenario, the model is expected to provide a prediction, even if it has not completed the entire inference process. The main focus here is ensuring the model can yield a usable (though potentially suboptimal) prediction at any moment, allowing for adaptive responses in real-time systems where the computation time is constrained or unpredictable. 

\emph{However, to our knowledge, current G2M methods ignore the urgent need for a more flexible and adaptive approach.} Our ProGMLP is designed to fill this gap by offering a progressive training and inference mechanism that enables dynamic adjustment of the trade-off between accuracy and inference cost. This makes ProGMLP particularly suitable for real-world applications where computational resources are limited, and the ability to control inference time is critical.

\subsection{Methodology}
\subsubsection{Notations}
Consider a graph $\mathcal{G} = \{\mathcal{V}, \mathcal{E}, Y\}$ where $\mathcal{V}$ and $\mathcal{E}$ represent the set of nodes and edges, respectively. Let $N$ denote the total number of nodes in the graph. The label matrix $Y \in \mathbb{R}^{N \times C}$ consists of one-hot vectors, where $C$ denotes the number of classes. The node features are represented by a matrix $X \in \mathbb{R}^{N \times d}$, where each row $x_{i}$ corresponds to the $d$-dimensional feature vector of node $i$. The adjacency matrix $A \in \mathbb{R}^{N \times N}$ encodes the graph's structure, with $A_{ij} = 1$ if there is an edge between nodes $i$ and $j$, and 0 otherwise. For the node classification task which involves predicting the labels of unlabeled nodes, we can divide the node set into labeled ($\labeledNodeSet$) and unlabeled ($\unLabeledNodeSet$) subsets, with corresponding feature matrices $X^{L}$ and $X^{U}$, and label matrices $Y^{L}$ and $Y^{U}$. In this paper, uppercase letters denote matrices, while lowercase letters represent specific rows within these matrices (e.g., $x_{i}$ is the $i$-th row of $X$). 
\begin{figure}
	\includegraphics[width=1\columnwidth]{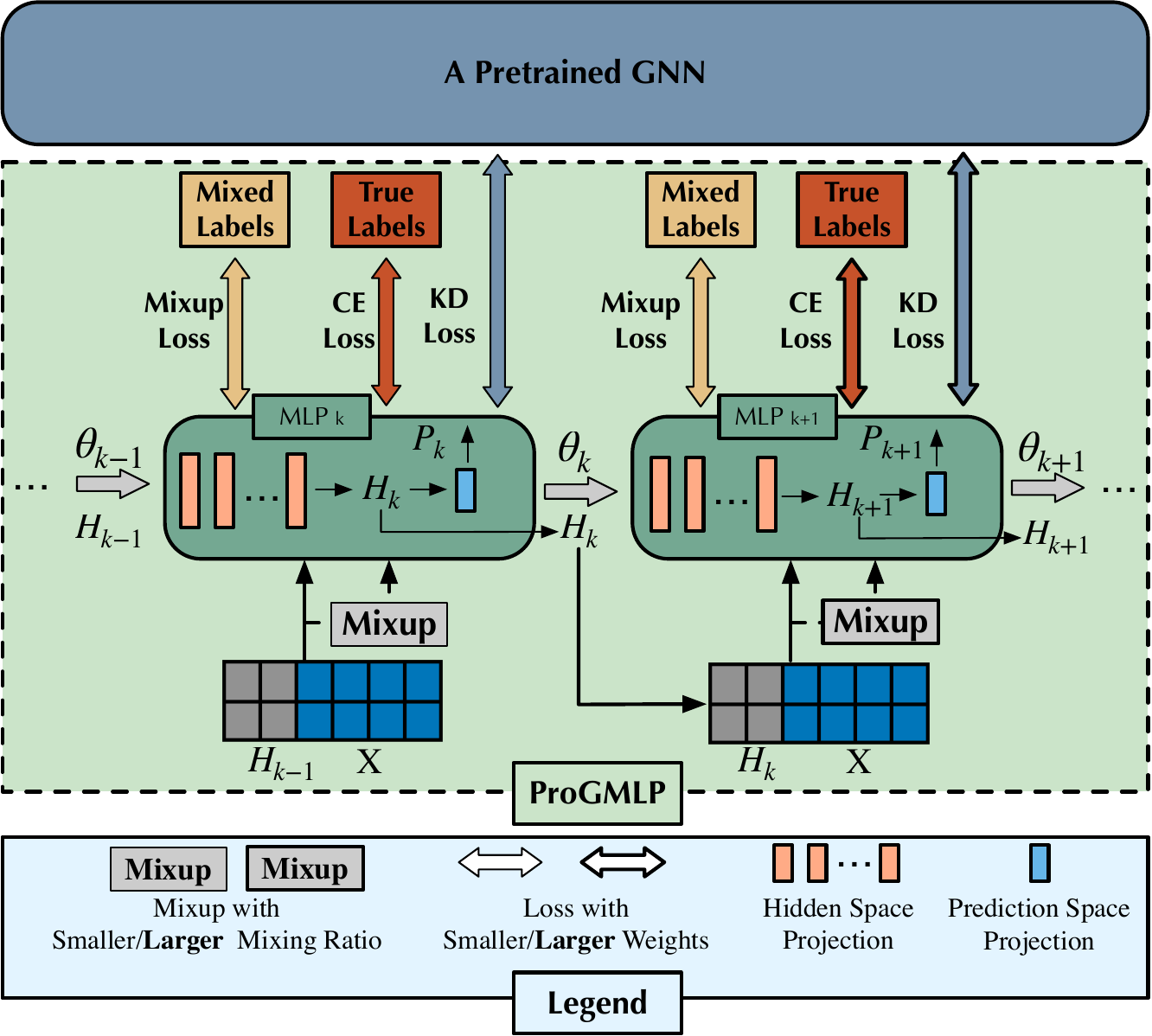}
	\caption{The training architecture of ProGMLP. \label{fig:progmlp}}
\end{figure}

\subsubsection{Overview}
ProGMLP is an ensemble framework designed to distill knowledge from a pre-trained GNN into a sequence of progressively trained MLPs, as illustrated in Figure~\ref{fig:progmlp}. The progressive training structure is established by initializing each student MLP $f_{k+1}$ with the parameters $\theta_k$ of the previously trained student $f_k$. A pre-trained GNN serves as the teacher, guiding the learning of this MLP sequence. For each student MLP $f_{k+1}$, the input comprises both the original node features $X$ and the hidden representation $H_k$ output by the previous MLP. The initial hidden representation $H_0$ is a zero matrix. Each MLP is trained using a combination of three loss functions: Mixup Loss (from interpolated samples and labels), Cross-Entropy Loss (from predictions and ground-truth labels), and Knowledge Distillation (KD) Loss (from the discrepancy between the teacher GNN and the student MLP outputs). This training process proceeds iteratively, with each subsequent MLP student refining the knowledge distilled by its predecessor. The mixup strategy employs an increasingly stronger mixing ratio over time, while the loss weights for KD and mixup components are progressively increased for later MLPs. This encourages deeper and more robust learning.

\subsection{Progressive Training Structure}
\label{sec:pro_structure}
The progressive training structure is at the heart of ProGMLP. Unlike conventional G2M methods that train a single student model, ProGMLP trains a sequence of student MLPs, each one initialized with the parameters of the previously trained model. This progressive approach ensures that the later models in the sequence start from a more advanced state and are capable of tackling more complex tasks. A $L$-layer MLPs student $f_{k}$, consisting of a ($L-1$)-layer fully connected network (FCNs) with the same hidden dimensionality for latent space projection and 1-layer FCN for prediction space projection, can be described as:
\begin{equation}
	H_{k}, P_{k} = f_{k}(X_{k}), \quad H_{k} \in \mathbb{R}^{N \times d^\prime}, P_{k} \in \mathbb{R}^{N \times C},
\end{equation}
where $X_{k} = \operatorname{CONCAT}(X, H_{k-1})$ is the input consisting of raw features $X$ and hidden representations $H_{k-1}$ from the previous student. Here, $H_{0} = \mathbf{O} \in \mathbb{R}^{N \times d^{\prime}}$ is a zero matrix. First, $f_{k}$ projects input $X_{k}$ into the hidden space within the first ($L-1$)-layer FCNs and then maps the hidden representations $H_{k}$ into the prediction space to obtain the predictions $P_{k}$ via the last FCN. The first MLPs student $f_{1}$ is trained with random initialization. Then, after training $f_{k}$, its parameters are used to initialize the next student $f_{k+1}$. The process is repeated for each subsequent student model, progressively refining the learning process. The relationship between these students can be expressed as:
\begin{equation}
	\theta_{k} \leftarrow \mathbf{Train}(f_{k-1} | \theta_{k-1}),
\end{equation}
where $\theta_{k-1}$ represents the parameters from the $(k-1)$-th student . For the $k$-th MLPs student, the parameters are optimized guided by the following loss function:
\begin{equation}
	\label{eq:loss_student_k}
	\mathcal{L}_{k} = \mathcal{L}^{PKD}_{k} + \mathcal{L}^{PMA}_{k},
\end{equation}
where $\mathcal{L}^{PKD}_{k}$ is the progressive knowledge distillation loss defined by Eq.~(\ref{eq:loss_pkd}) and $\mathcal{L}^{PMA}_{k}$ is the progressive mixup loss defined by Eq.~(\ref{eq:loss_pma}). We will introduce them in the following sections.

The idea behind this approach is to incrementally build on the learned knowledge, much like how a Recurrent Neural Network (RNN) propagates hidden states through time steps. By initializing each student model with the parameters of its predecessor, ProGMLP allows the knowledge to be gradually refined, leading to models that are both more accurate and more capable of handling diverse inputs.

\subsection{Progressive Knowledge Distillation}
\label{sec:pro_kd}
Progressive Knowledge Distillation is a combination of the cross-entropy loss and the Kullback-Leibler (KL) divergence~\cite{kd_kl,kd_l2}, allowing each student model to refine its predictions based on both the true labels and the predictions of the previous model. In this context, each student in the sequence is expected to improve upon the predictions of the previous one. The loss function for student $k$ can be defined as:
\begin{equation}
	\label{eq:loss_pkd}
	\begin{aligned}
	\mathcal{L}^{PKD}_{k} &= k^{\beta}\left(\alpha \mathcal{L}^{CE}_{k} + (1 - \alpha) \mathcal{L}^{KD}_{k}\right) \\
					&= k^{\beta}( \frac{\alpha}{|\labeledNodeSet|} \sum_{i \in \labeledNodeSet} \ell^{CE} \left(f_{k}(x_{i}), y_{i}\right) \\ 
					&+\frac{1-\alpha}{{|\mathcal{V}|}} \sum_{i \in \mathcal{V}} \ell^{KD} \left(z^{g}_{i}, f_{k}(x_{i}) \right) ),
	\end{aligned}
\end{equation}
where $\ell^{CE}$ is the cross-entropy loss between the predictions and true labels and $\ell^{KD}$ is the KL divergence between the current student predictions $f_{k}(x)$ and the GNNs' predictions $z^{g}$. $\beta$ is a factor that increases the weight of the loss for later student in the sequence, encouraging more accurate predictions and $\alpha$ is the weighting factor that balances the distillation loss and the standard cross-entropy loss. The loss function is applied to each student output, but with different weights $k^{\beta}$. The later students in the sequence are assigned higher weights, incentivizing them to make more accurate predictions.

\subsection{Progressive Mixup Augmentation}
\label{sec:pro_mixup}
The Progressive Mixup Augmentation in ProGMLP is designed to gradually increase the difficulty of the learning tasks assigned to each student model. Mixup is a data augmentation technique that generates new training examples by linearly interpolating between pairs of examples. In ProGMLP, we produce the progressive hard examples for student $k \geq 1$ as follows:
\begin{equation}
	\label{eq:mixup}
	\begin{aligned}
		&x_{ij} = \lambda \operatorname{CONCAT}(x_{i}, H_{k-1}[i,:]) \\
		&\qquad+ (1 - \lambda) \operatorname{CONCAT}(x_{j}, H_{k-1}[j,:]) \\
		&y_{ij} = \lambda y_{i} + (1 - \lambda) y_{j},
	\end{aligned}
\end{equation}
where $\lambda$ is the mixing rate and $H_{k-1}[i,:]$ is the $i$-th row vector of hidden representations $H_{k-1}$. Then, the progressive mixup loss $\mathcal{L}^{PMA}$ for mixed samples is:
\begin{equation}
	\label{eq:loss_pma}
	\mathcal{L}^{PMA}_{k} = \frac{1}{|\labeledNodeSet|} \sum_{(i,j) \in \mathcal{D}^{m}_{k}} \ell^{CE} \left(f_{k}(x_{ij}), y_{ij}\right),
\end{equation}
where $\mathcal{D}^{m}_{k}$ is the mixup pair set for the $k$-th student MLPs. 

From Eq.~(\ref{eq:mixup}), \emph{we can see that a $\lambda$ close to 0 results in examples that are almost identical to one of the original examples, while a $\lambda$ close to 0.5 results in examples that are more challenging and significantly different from either of the original examples.} By adjusting $\lambda$ over time, we can control the difficulty of the examples presented to each MLPs student in the ProGMLP sequence:
\begin{equation}
	\label{eq:lambda_adjustment}
	\lambda_{k} \leftarrow \min \left(\max \left(\lambda_{k-1} + \gamma(\bar{\ell} - \tau), 0\right), 0.5\right)
\end{equation}
Here, $\lambda_{k}$ is the mixing factor for the $k$-th MLPs student, $\gamma > 0$ is an adjustment rate which determines how sensitively $\lambda$ responds to changes in the model's learning process, $\bar{\ell}$ is the moving average of the mixup loss for the current student, and $\tau$ is a predefined threshold that acts as a reference point for adjusting $\lambda$. When the moving average loss is close to $\tau$, $\lambda$ remains stable; when the loss decreases (indicating better performance), $\lambda$ increases. In this paper, we set $\tau=0.1$.

$\bar{\ell}$ is typically calculated using an exponential moving average (EMA), which gives more weight to recent losses while still considering past losses. The formula for calculating the exponential moving average at time step $t$ (iteration) is:
\begin{equation}
	\label{eq:avg_loss}
	\bar{\ell}_{t} = \sigma \bar{\ell}_{t-1} + (1 - \sigma) \bar{\ell}_{t},
\end{equation} 
where $\bar{\ell}_{t}$ is the moving average loss at the iteration $t$ and $\sigma$ is a smoothing factor between 0 and 1, controlling how much weight is given to past losses. A higher $\sigma$ value means $\bar{\ell}$ changes more slowly, giving more weight to past losses. $\bar{\ell}_{0}$ is initialized at the current mixup loss calculated by Eq.~(\ref{eq:loss_pkd}). In this paper, we set $\sigma=0.1$.

As the student models become more capable, they are exposed to more challenging examples, which pushes them to learn more complex patterns and generalize better. This approach aligns with the progressive nature of ProGMLP, where each model is expected to handle more difficult tasks than its predecessor.

\begin{algorithm}
\caption{Training Algorithm for ProGMLP}
\label{alg:progmlp}
\begin{algorithmic}[1]
\Require Pretrained GNN teacher $GNN_T$, node features $X$, number of MLP students $K$, number of epochs $E$
\State \textbf{Initialize:} $H_0 \gets \mathbf{O}$, parameters of $f_1$: $\theta_1$
\State Compute teacher outputs: $z^{g} \gets g(X, A)$
\For{$k = 1$ to $K$} 
	\Comment{Training each student}
    \For{epoch = 1 to $E$}
        \State $(H_k, P_k) \gets f_k(\text{CONCAT}(X, H_{k-1}))$        
        \State Compute the PKD loss via Eq.~(\ref{eq:loss_pkd})
        \State Generate mixed examples $(x_{ij}, y_{ij})_{(i,j) \in \mathcal{D}^m_k}$ via Eq.~(\ref{eq:mixup})
        \State Compute the PMA loss via Eq.~(\ref{eq:loss_pma})
        \State Compute total loss $\mathcal{L}_k$ combining Eq.~(\ref{eq:loss_pkd}) and Eq.~(\ref{eq:loss_pma})
        \State Update parameters: $\theta_k \gets \theta_k - \eta \nabla_{\theta_k} \mathcal{L}_k$
    \EndFor
    \State Initialize student $f_{k+1}$ with parameters $\theta_k$
    \State Pass hidden state $H_k$ to next student $f_{k+1}$
\EndFor
\State \Return Trained MLP students $\{f_1, \dots, f_K\}$
\end{algorithmic}
\end{algorithm}

\subsection{Train and Inference}
\label{sec:train_inference}
\subsubsection{Training}
ProGMLP consists of $K$ student MLPs, where the $k$-th student $f_{k}$ is trained using the loss function $\mathcal{L}_{k}$, as defined in Eq.~(\ref{eq:loss_student_k}). Each student is trained for up to $E_{1}$ epochs, with early stopping applied based on a patience criterion of $E_{2}$ epochs, where $E_{2} < E_{1}$. This ensures that the training process is efficient and prevents overfitting. The training algorithm is provided in Alg.~\ref{alg:progmlp}.

\subsubsection{Inference}
During inference, ProGMLP evaluates each student model sequentially. The inference process is designed to stop once certain conditions are met, ensuring a balance between efficiency and accuracy. One of the primary criteria for halting the inference process is the prediction confidence of the current student model.

The process halts when the confidence of the $k$-th student, denoted as $c_{k}$, exceeds a predefined threshold $\tau^{conf}$. This threshold reflects the minimum confidence level required for the system to consider the prediction sufficiently accurate. The prediction confidence $c_{k}$ for the $k$-th student is computed as the average of the maximum softmax probabilities across all unlabeled nodes: 
\begin{equation}
	\label{eq:conf}
	c_{k} = \underset{i \in \unLabeledNodeSet}{\text{mean}} \left({\underset{j \in {1, ..., C}}{\max} \operatorname{SOFTMAX}\left(f_{k}(x_{i})\right)_{j}}\right),
\end{equation}
where $\operatorname{SOFTMAX}(f_{k}(x_{i}))$ converts the logits for sample $i$ into a probability distribution over classes and $\operatorname{SOFTMAX}(f_{k}(x_{i}))_{j}$ is the predicted probability on $j$-th class for sample $i$.

If the confidence threshold is not met, ProGMLP continues to evaluate the next student model $f_{k+1}$. The model stops at the first student where $c_{k} \geq \tau^{conf}$, ensuring efficient inference by avoiding unnecessary evaluations.

Assume $ \{P_{1}, P_{2}, \cdots, P_{k}\}$ are the outputs from $k$ executed students with corresponding confidences $\{c_{1}, c_{2}, \cdots, c_{k}\}$. The final prediction $P$ is computed as a confidence-weighted sum of the predictions from the evaluated models:
\begin{equation}
	\label{eq:pred}
	P = \sum_{j=1}^{k} w_{j}P_{j},
\end{equation}
where $w_{k} = \operatorname{SOFTMAX}(\{c_{1}, c_{2}, \cdots, c_{k}\})_{k}$ is the normalized weight calculated from model confidence. This approach ensures that ProGMLP maintains flexibility and efficiency, making it suitable for a wide range of runtime conditions and application requirements.

\subsection{Complexity}
The time complexity of the ProGMLP framework is primarily determined by the number of MLPs students $K$. The time complexity of training each MLPs student mainly comes from: (1) feature forward $O(Nd^\prime(d + d^{\prime} + C))$; (2) knowledge distillation $O(NC)$; (3) mixup augmentation $O(|\labeledNodeSet|(d + d^{\prime}))$, where $|\labeledNodeSet|$ and $d^{\prime}$ are largely small compared to $N$ and $d$, respectively. Therefore, the overall training complexity of ProGMLP can be approximated as $O\left(KN(d^\prime(d + d^{\prime} + C) +C)\right)$. During inference, ProGMLP evaluates each MLPs student sequentially, stopping the process based on the confidence-time budgeting mechanism. \emph{In the worst-case scenario}, all $K$ MLPs are evaluated. The inference time complexity for a single sample is therefore $O\left(Kd^\prime(d + d^{\prime} + C)\right)$. 

\section{Experiments}

\begin{table*}[!htbp]
\begin{center}
\caption{Datasets Statics.\label{tab:data_stat}}
\renewcommand\arraystretch{1.2}
\begin{tabular}{lcccc}
\toprule
\textbf{Dataset} & \textbf{\# Nodes} & \textbf{\# Edges} & \textbf{\# Features} & \textbf{\# Classes}  \\ \midrule
Cora & 2,708 & 5,278 & 1,433 & 7  \\
Pubmed & 19,717 & 44,324 & 500 & 3  \\
Amazon Photo & 7,650 & 119,081 & 745 & 8  \\
Amazon Computers & 13,381 & 245,778 & 767 & 10 \\
Coauthor CS & 18,333 & 81,894 & 6,805 & 15  \\
Coauthor Physics & 34,493 & 247,962 & 8,415 & 5  \\
ogbn-arxiv & 169,343 & 1,166,243 & 128 & 40 \\ 
ogbn-products & 2,449,029 & 61,859,140 & 128 & 47\\
\bottomrule
\end{tabular}
\end{center}
\end{table*}

In this section, we present a comprehensive set of experiments to evaluate the effectiveness of ProGMLP\footnote{https://github.com/WeigangLu/ProGMLP}, including its ability to balance the trade-off between performance and inference cost (cf. Sec.~\ref{sec:tradeoff_exp}), performance on the transductive node classification task using medium-scale graphs (cf. Sec.~\ref{sec:medium_scale}) and large scale-graphs (cf. Sec.~\ref{sec:large_scale}), and performance on the inductive node classification task (cf. Sec.~\ref{sec:ind}). Besides, we conduct extensive experiments to investigate the role of each component within ProGMLP (cf. Sec.~\ref{sec:ablation_study}) and the analysis of hyperparameters (cf. Sec.~\ref{sec:hyper_analysis}). 


\subsection{Hardware and Software}
ProGMLP is implemented based on the Torch Geometric library~\cite{torch_geo} and PyTorch 3.7.1 with Intel(R) Core(TM) i9-10980XE CPU @ 3.00GHz and one NVIDIA A100 GPUs with 40GB memory.

\subsection{Datasets} 
To comprehensively evaluate the performance, generalizability, and scalability of ProGMLP, we have selected \emph{eight} widely-adopted real-world graph datasets: 
\begin{itemize}
	\item \textbf{Cora}~\cite{cora}: This is a widely-used citation network dataset where nodes represent academic publications and edges represent citation links. Each publication is described by a binary word vector indicating the absence or presence of words from a dictionary. The task is to classify each document into one of seven research categories.
	\item \textbf{Pubmed}~\cite{cora}: Similar to Cora, Pubmed is another standard citation network benchmark, but larger in scale. 
	\item \textbf{Amazon Photo}~\cite{coauthor}: This dataset is a co-purchase network where nodes represent products sold on Amazon and edges connect items that are frequently bought together. Node features are derived from product reviews, and the task is to classify products into their respective categories. 
	\item \textbf{Amazon Computers}~\cite{coauthor}: As another co-purchase graph from Amazon, this dataset is larger and denser than Amazon Photo. Nodes represent computers and related hardware, with edges indicating co-purchasing behavior. 
	\item \textbf{Coauthor CS}~\cite{coauthor}: This dataset represents a co-authorship network from the field of computer science. Nodes are authors, who are connected by an edge if they have co-authored a paper, and node features are based on keywords from their publications.
	\item \textbf{Coauthor Physics}~\cite{coauthor}: Similar to Coauthor CS, this is a co-authorship graph, but it focuses on authors from the physics domain and is larger in scale. 
	\item \textbf{ogbn-arxiv}~\cite{ogb}: This is a large-scale citation graph from the Open Graph Benchmark (OGB) that represents a directed network of all Computer Science arXiv papers. 
	\item \textbf{ogbn-products}~\cite{ogb}: This is a massive co-purchasing network, also from the OGB, where nodes represent products sold on Amazon. 
\end{itemize}
These datasets exhibit diversity in node features, graph structures, and task complexities, offering a comprehensive benchmark for evaluating the generalizability of our approach. The details of all the graphs are provided in Table~\ref{tab:data_stat}.

\subsection{Teacher Models} 

For the teacher GNN models, we select three of the most representative architectures: 
\begin{itemize}
	\item \textbf{GCN}~\cite{gcn} introduces a convolutional neural network architecture for graphs by aggregating feature information from local neighborhoods.
	\item \textbf{GAT}~\cite{gat} incorporates attention mechanisms into graph neural networks, allowing nodes to weigh the importance of their neighbors' features during the aggregation process. 
	\item \textbf{GraphSAGE}~\cite{sage} is an inductive framework that generates node embeddings by sampling and aggregating features from a node's local neighborhood, enabling scalable learning on large graphs.
\end{itemize}
Each GNN is implemented with a standard 2-layer structure to ensure a fair and consistent comparison across experiments.

\subsection{Student Models} 

\textbf{Ensemble G2M Methods.} Given the requirements of early exit and anytime inference settings where a model must be capable of producing predictions even if its execution is interrupted. However, standard G2M methods, which only provide output upon complete execution. Thus, we compare ProGMLP against two representative G2M methods, GLNN~\cite{glnn} and AdaGMLP~\cite{adagmlp}, with the following modifications to suit the early exit and anytime inference scenarios:

\begin{itemize} 
	\item \textbf{E-GLNN$_{K}$}: This is a $K$-ensemble version of GLNN~\cite{glnn}, a pioneering G2M method. E-GLNN$_{K}$ consists of $K$ student blocks, each block being a 2-layer MLPs. If the model is interrupted or exits early, predictions from the $k$ completed students $\{1, 2, \cdots, k\}$ are averaged and output as the final prediction. When $K = 1$, E-GLNN is reduced to the standard one-student architecture.
	
	\item \textbf{AdaGMLP$_{K}$}: An ensemble G2M framework~\cite{adagmlp} that employs the AdaBoost algorithm for training and inference. Similar to E-GLNN, the final prediction in AdaGMLP is the weighted sum of the predictions from the $k$ already-executed student models, with the weights re-normalized according to the AdaBoosting algorithm. 
\end{itemize} 

\begin{figure*}[!htbp]
\begin{center}
	\subfigure[Teacher = GCN]{
			\includegraphics[width=0.9\linewidth]{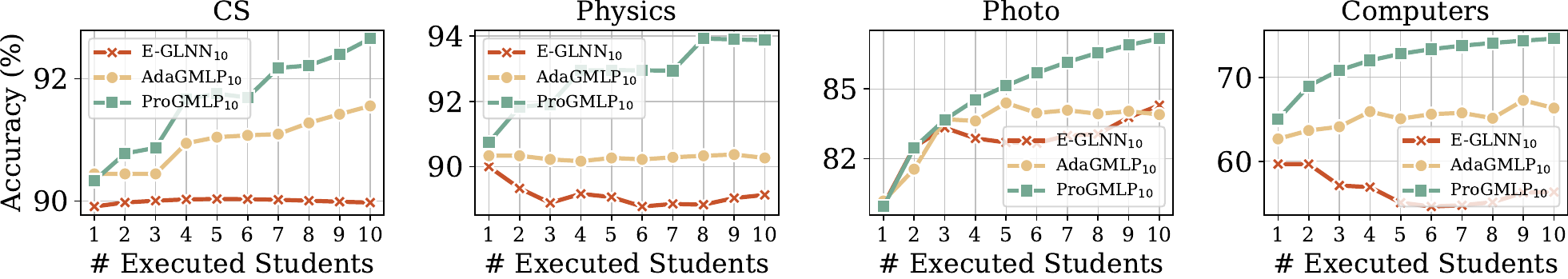}
	}
	\subfigure[Teacher = GraphSAGE]{
			\includegraphics[width=0.9\linewidth]{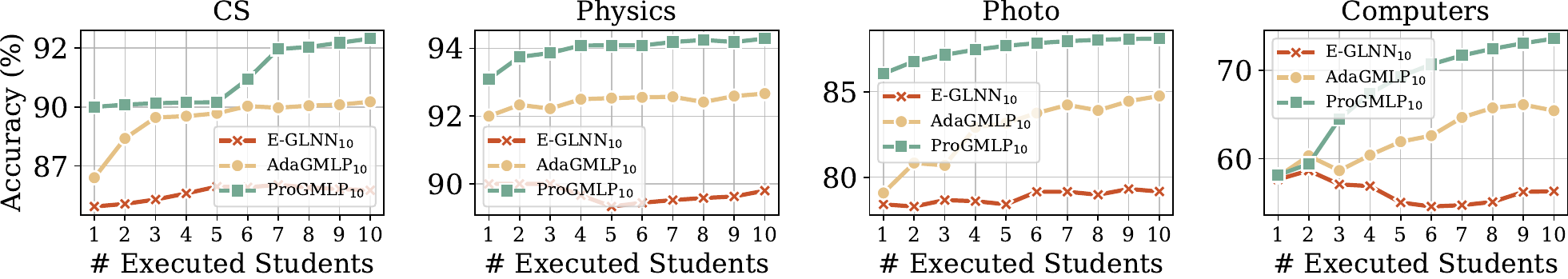}
	}

	\caption{Accuracy vs. Inference Cost (\# Number of Executed Students) for ProGMLP, AdaGMLP, and GLNN. Each curve represents the progression of accuracy as more student models are executed, highlighting the trade-off between performance and inference cost. ProGMLP consistently outperforms the baseline methods, particularly as more student models are executed, achieving higher accuracy with a better balance between computational efficiency and performance. \label{fig:tradeoff}}
\end{center}
\end{figure*}

\textbf{Non-Ensemble G2M Methods.} Additionally, we choose three state-of-the-art (SOTA) non-ensemble G2M methods:
\begin{itemize}
	\item \textbf{NOSMOG}~\cite{nosmog}: It aims to enhance the MLP student's performance by incorporating structural information from the graph. 
	\item \textbf{KRD}~\cite{krd}: It focuses on improving the quality of the knowledge transferred from the GNN teacher to the MLP student. It achieves this by introducing a reliable sampling strategy to select more informative instances for the distillation process.
	\item \textbf{HGMD}~\cite{hgmd}: It decouples and estimates two types of distillation hardness—knowledge hardness and distillation hardness—to better transfer knowledge from GNNs to MLPs.
\end{itemize}
%
%

\subsection{Parameters}
The parameter search space for all the methods (including teachers and students) is listed as follows:
\begin{itemize}
	\item learning rate: $\{0.001, 0.002, 0.005, 0.01, 0.02, 0.05\}$
	\item weight decay rate: $\{5e-4, 5e-5, 5e-7, 5e-9\}$
	\item hidden dimensionality: $\{16, 32, 64, 128, 256, 512\}$
	\item dropout: $\{0.1, 0.2, \cdots, 0.9\}$
	\item teacher model depth: $\{2, 3\}$
	\item number of students: $\{2, 3, \cdots, 6\}$
\end{itemize}
Besides, we apply early stopping with patience for 50 epochs on each trail.

\subsection{Trade-off Ability Evaluation}
\label{sec:tradeoff_exp}
In this section, we evaluate the effectiveness of ProGMLP in balancing the trade-off between model performance (in terms of accuracy) and inference cost (measured by the number of executed student models). In real-world applications, particularly those involving resource-constrained environments, achieving high accuracy with minimal computational cost is crucial. ProGMLP’s progressive training and inference structure is designed to enable dynamic adjustment of this trade-off. For all the G2M methods, we set $K = 10$ and use the same configuration for all the MLPs students by setting hidden dimensionality at 128, dropout rate at 0.5, learning rate at 0.001, and weight decay rate at 0.0005. \emph{During the inference stage, we stop them at the $k$-th student from 1 to 10 to understand how each method improves accuracy with increasing computational resources.}

The results, as shown in Figure~\ref{fig:tradeoff}, indicate that ProGMLP provides substantial early gains in accuracy, achieving competitive performance with only a few executed student models. This is particularly important for scenarios with strict computational constraints. As more students are executed, ProGMLP continues to show steady accuracy improvement, outperforms both GLNN and AdaGMLP. GLNN, in contrast, plateaus after a few students, while AdaGMLP shows slower, incremental improvements. The progressive learning structure of ProGMLP allows each subsequent student model to refine its predictions, contributing to its superior performance across the board.

Overall, ProGMLP strikes a better balance between performance and inference cost compared to the baselines. It reaches near-optimal accuracy with fewer student models, making it more efficient for a variety of runtime scenarios. The experiment highlights that ProGMLP is highly adaptable, capable of delivering high accuracy quickly when needed, but also capable of continuing to improve when computational resources are available, making it ideal for diverse real-world applications.

\begin{table*}[!htbp]
\begin{center}
\caption{Comparison with Ensemble G2M Methods.\label{tab:semi}}

\renewcommand\arraystretch{1.4}
\begin{tabular}{l cccccc c}

\toprule
\textbf{Teacher}  & \multirow{2}{*}{Cora} & \multirow{2}{*}{Pubmed}  & \multirow{2}{*}{CS} & \multirow{2}{*}{Physics} & \multirow{2}{*}{Photo} & \multirow{2}{*}{Computers} & \multirow{2}{*}{\textbf{\emph{Impro.}}} \\ 
~+\textbf{Student} & & & & & &\\
\midrule

GCN & $79.03_{\pm0.37}$ & 76.66$_{\pm0.35}$ & 90.68$_{\pm0.17}$ & 93.59$_{\pm0.59}$ & 88.57$_{\pm0.83}$ & 77.82$_{\pm0.88}$ & 0.00\% \\
\hdashline[3pt/3pt]

~+E-GLNN$_{2}$ & 79.28$_{\pm0.62}$ & 76.60$_{\pm0.48}$ & 89.88$_{\pm0.16}$ & 92.11$_{\pm0.82}$ & 84.42$_{\pm1.18}$ & 75.39$_{\pm1.17}$ & \cellcolor[HTML]{FDECEE}-1.67\% \\
~+E-GLNN$_{4}$ & 79.30$_{\pm0.83}$ & 76.98$_{\pm0.53}$ & 89.79$_{\pm0.19}$ & 92.65$_{\pm0.33}$ & 84.73$_{\pm1.64}$ & 75.82$_{\pm1.12}$ & \cellcolor[HTML]{FDECEE}-1.36\% \\

~+AdaGMLP$_{2}$ & 79.15$_{\pm0.19}$ & 76.84$_{\pm0.64}$ & 90.53$_{\pm0.12}$ & 92.58$_{\pm0.26}$ & 84.49$_{\pm1.53}$ & 76.34$_{\pm1.26}$ & \cellcolor[HTML]{FDECEE}-1.23\% \\
~+AdaGMLP$_{4}$ & 79.36$_{\pm1.12}$ & 77.21$_{\pm0.66}$ & 91.69$_{\pm0.37}$ & 92.71$_{\pm0.29}$ & 85.26$_{\pm1.61}$ & 77.11$_{\pm1.19}$ & \cellcolor[HTML]{FDECEE}-0.56\% \\

\rowcolor[HTML]{E5F2FC}~+\textbf{ProGMLP} & \textbf{80.19}$_{\pm0.39}$ & \textbf{77.42}$_{\pm0.35}$ & \textbf{92.43}$_{\pm0.12}$ & \textbf{94.23}$_{\pm0.04}$ & \textbf{90.91}$_{\pm0.77}$ & \textbf{78.82}$_{\pm0.44}$ & \cellcolor[HTML]{FDECEE}\textbf{+1.50\%} \\
\midrule

GAT & 78.47$_{\pm2.77}$ & 75.71$_{\pm0.46}$ & 90.42$_{\pm0.66}$ & 92.88$_{\pm0.24}$ & 86.48$_{\pm1.51}$ & 76.82$_{\pm1.23}$ & 0.00\% \\
\hdashline[3pt/3pt]

~+E-GLNN$_{2}$ & 78.94$_{\pm2.05}$ & 76.84$_{\pm0.76}$ & 89.72$_{\pm0.70}$ & 90.63$_{\pm0.95}$ & 82.66$_{\pm1.69}$ & 74.78$_{\pm2.16}$ & \cellcolor[HTML]{FDECEE}-1.36\% \\
~+E-GLNN$_{4}$ & 79.04$_{\pm2.12}$ & 77.08$_{\pm0.44}$ & 90.66$_{\pm0.41}$ & 90.98$_{\pm0.48}$ & 83.01$_{\pm1.63}$ & 74.54$_{\pm2.34}$ & \cellcolor[HTML]{FDECEE}-1.04\% \\

~+AdaGMLP$_{2}$ & 78.95$_{\pm2.44}$ & 76.92$_{\pm0.79}$ & 90.42$_{\pm0.25}$ & 92.94$_{\pm0.18}$ & 84.39$_{\pm1.86}$ & 76.85$_{\pm1.17}$ & \cellcolor[HTML]{FDECEE}-0.02\% \\
~+AdaGMLP$_{4}$ & 79.95$_{\pm2.50}$ & 77.25$_{\pm0.46}$ & 90.77$_{\pm0.14}$ & 93.07$_{\pm0.29}$ & 85.64$_{\pm1.69}$ & 77.31$_{\pm1.38}$ & \cellcolor[HTML]{FDECEE}+0.70\% \\

\rowcolor[HTML]{E5F2FC}~+\textbf{ProGMLP}  & \textbf{80.50}$_{\pm2.08}$ & \textbf{77.63}$_{\pm0.82}$ & \textbf{91.83}$_{\pm0.44}$ & \textbf{94.16}$_{\pm0.14}$ & \textbf{88.94}$_{\pm1.64}$ & \textbf{79.16}$_{\pm1.29}$ & \cellcolor[HTML]{FDECEE}\textbf{+2.33\%} \\
\midrule

GraphSAGE & 78.56$_{\pm0.64}$ & 75.39$_{\pm0.49}$ & 91.84$_{\pm0.46}$ & 92.37$_{\pm1.44}$ & 86.54$_{\pm0.69}$ & 79.32$_{\pm0.31}$ & 0.00\% \\
\hdashline[3pt/3pt]
~+E-GLNN$_{2}$ & 78.37$_{\pm0.80}$ & 76.09$_{\pm0.65}$ & 91.11$_{\pm0.13}$ & 92.24$_{\pm0.41}$ & 84.55$_{\pm1.16}$ & 76.28$_{\pm0.29}$ & \cellcolor[HTML]{FDECEE}-1.06\% \\
~+E-GLNN$_{4}$ & 78.32$_{\pm0.97}$ & 76.95$_{\pm0.79}$ & 90.62$_{\pm0.36}$ & 92.41$_{\pm0.39}$ & 84.78$_{\pm2.38}$ & 76.14$_{\pm0.32}$ & \cellcolor[HTML]{FDECEE}-0.93\% \\

~+AdaGMLP$_{2}$  & 78.39$_{\pm0.76}$ & 77.01$_{\pm0.84}$ & 92.41$_{\pm0.20}$ & 92.67$_{\pm0.29}$ & 85.16$_{\pm0.84}$ & 77.94$_{\pm1.41}$ & \cellcolor[HTML]{FDECEE}-0.08\% \\
~+AdaGMLP$_{4}$  & 78.84$_{\pm0.98}$ & 77.31$_{\pm0.29}$ & 92.79$_{\pm0.21}$ & \textbf{94.20}$_{\pm0.30}$ & 86.98$_{\pm0.72}$ & 79.23$_{\pm1.06}$ & \cellcolor[HTML]{FDECEE}+1.05\% \\
\rowcolor[HTML]{E5F2FC}~+\textbf{ProGMLP} & \textbf{79.28}$_{\pm0.86}$ & \textbf{77.84}$_{\pm1.05}$ & \textbf{93.13}$_{\pm0.10}$ & 93.70$_{\pm0.16}$ & \textbf{87.48}$_{\pm1.74}$ & \textbf{80.21}$_{\pm1.21}$ & \cellcolor[HTML]{FDECEE}\textbf{+1.54\%} \\

\bottomrule 
\end{tabular}
\end{center}
\end{table*}

\subsection{Comparison with Ensemble G2M Methods}
\label{sec:medium_scale}
In this section, we compare our ProGMLP with ensemble G2M methods in the trandustive node classification task across 6 medium-scale graphs with standard splits according to their original settings. Each result is obtained from \emph{10} different runs. For the ease of parameter search, we set $\alpha=0.5$, $\beta=0.8$, $\gamma=0.9$, and $\tau^{conf} = 0.9$. 

ProGMLP consistently outperforms both GLNN and AdaGMLP across almost all datasets, demonstrating its strength in balancing accuracy and inference efficiency. One of the key reasons for ProGMLP’s superior performance is its progressive training structure, which allows each MLP student to progressively refine its learning from previous models, leading to more accurate predictions. For instance, on the CS and Photo datasets, ProGMLP shows substantial improvements over both GLNN and AdaGMLP, highlighting its ability to effectively generalize across various datasets and handle different levels of complexity in node classification tasks.

GLNN, on the other hand, tends to lag behind both ProGMLP and AdaGMLP in terms of accuracy, especially as the complexity of the dataset increases. While GLNN provides faster inference due to its simpler structure, it struggles to maintain the high accuracy levels seen in ProGMLP, especially in datasets like Photo and Computers. AdaGMLP, while generally more accurate than GLNN, also falls short when compared to ProGMLP. AdaGMLP’s use of an ensemble of MLP students improves performance over GLNN, but it lacks the dynamic and progressive learning enhancements of ProGMLP. As a result, AdaGMLP shows competitive performance but does not reach the same accuracy improvements, particularly in datasets like Physics and CS, where ProGMLP’s advanced training techniques lead to more refined predictions.

\subsection{Comparison with Non-Ensemble G2M Methods}
\label{sec:non_ensemble}

To further evaluate the performance of ProGMLP, we also conduct experiments comparing its final performance (i.e., after all student MLPs are potentially executed or the confidence threshold is met) against several state-of-the-art GNN-to-MLP (G2M) distillation methods that primarily train a single student MLP. For a fair comparison in this setting, we consider the output of ProGMLP when it has utilized its full sequence of students or achieved its early-exit confidence criterion, representing its best achievable accuracy. We compared ProGMLP with KRD, HGMD, and NOSMOG, using GCN, GAT, and GraphSAGE as teacher models. The results are presented in Table~\ref{tab:non_ensemble}.

The empirical results demonstrate that ProGMLP consistently achieves highly competitive or superior accuracy. When GCN serves as the teacher, ProGMLP obtains the highest scores on the Physics (94.23\%) and Computers (78.82\%) datasets, while remaining closely competitive with KRD on CS and Photo. Notably, under the GAT teacher, ProGMLP uniformly surpasses all other G2M student models across all four datasets. 

These findings offer valuable insights into ProGMLP's capabilities. The strong performance, even when compared to methods specifically designed to optimize a single MLP student, underscores the power of ProGMLP. This indicates that the progressive refinement of knowledge and adaptive learning inherent in our framework not only facilitate flexible inference but also culminate in a powerful final model. In summary, ProGMLP proves to be a robust framework that excels not only in providing dynamic accuracy-efficiency trade-offs but also in achieving high peak predictive accuracy against specialized single-student G2M distillation techniques.

\begin{table}[!htbp]
\begin{center}
\caption{Comparison with Non-Ensemble G2M Methods.\label{tab:non_ensemble}}

\renewcommand\arraystretch{1.4}

\begin{tabular}{l cccc}

\toprule
\textbf{Teacher}  & \multirow{2}{*}{CS} & \multirow{2}{*}{Physics} & \multirow{2}{*}{Photo} & \multirow{2}{*}{Computers} \\ 
~+\textbf{Student} & & & & \\

\midrule

GCN & 90.68$_{\pm0.17}$ & 93.59$_{\pm0.59}$ & 88.57$_{\pm0.83}$ & 77.82$_{\pm0.88}$ \\
\hdashline[3pt/3pt]

~+KRD & \textbf{93.84}$_{\pm0.89}$ & 94.09$_{\pm1.21}$ & \textbf{91.18}$_{\pm0.80}$ & 78.18$_{\pm1.07}$ \\
~+HGMD & 92.00$_{\pm1.24}$	& 93.56$_{\pm1.43}$ & 90.49$_{\pm0.92}$	& 78.00$_{\pm1.36}$ \\
~+NOSMOG & 93.42$_{\pm0.95}$ & 93.75$_{\pm1.12}$ & 90.42$_{\pm0.84}$ & 78.81$_{\pm1.25}$ \\

\rowcolor[HTML]{E5F2FC}~+\textbf{ProGMLP} & 92.43$_{\pm0.12}$ & \textbf{94.23}$_{\pm0.04}$ & 90.91$_{\pm0.77}$ & \textbf{78.82}$_{\pm0.44}$ \\

\midrule

GAT & 90.42$_{\pm0.66}$ & 92.88$_{\pm0.24}$ & 86.48$_{\pm1.51}$ & 76.82$_{\pm1.23}$ \\
\hdashline[3pt/3pt]

~+KRD & 91.69$_{\pm1.24}$ & 93.64$_{\pm0.78}$ & 88.62$_{\pm1.96}$ & 78.60$_{\pm1.82}$\\
~+HGMD & 86.55$_{\pm3.26}$ & 91.66$_{\pm2.85}$ & 74.96$_{\pm2.33}$ & 75.65$_{\pm4.11}$\\
~+NOSMOG & 91.46$_{\pm1.03}$ & 93.28$_{\pm0.96}$ & 88.73$_{\pm1.82}$ & 78.04$_{\pm1.54}$\\

\rowcolor[HTML]{E5F2FC}~+\textbf{ProGMLP} & \textbf{91.83}$_{\pm0.44}$ & \textbf{94.16}$_{\pm0.14}$ & \textbf{88.94}$_{\pm1.64}$ & \textbf{79.16}$_{\pm1.29}$ \\
\midrule

GraphSAGE & 91.84$_{\pm0.46}$ & 92.37$_{\pm1.44}$ & 86.54$_{\pm0.69}$ & 79.32$_{\pm0.31}$ \\
\hdashline[3pt/3pt]

~+KRD & \textbf{93.64}$_{\pm0.98}$ & 93.40$_{\pm2.10}$ & 87.47$_{\pm0.82}$ & 80.01$_{\pm1.06}$\\
~+HGMD & 90.23$_{\pm1.42}$ & 91.15$_{\pm2.63}$ & 87.15$_{\pm1.14}$ & 79.33$_{\pm1.32}$\\
~+NOSMOG & 91.16$_{\pm0.84}$ & 93.22$_{\pm1.96}$ & 86.92$_{\pm0.86}$ & 79.56$_{\pm1.02}$\\

\rowcolor[HTML]{E5F2FC}~+\textbf{ProGMLP} & 93.13$_{\pm0.10}$ & \textbf{93.70}$_{\pm0.16}$ & \textbf{87.48}$_{\pm1.74}$ & \textbf{80.21}$_{\pm1.21}$ \\

\bottomrule 
\end{tabular}
\end{center}
\end{table}

\subsection{Evaluation on Large-scale Graphs}
\label{sec:large_scale}
\begin{table}[!th]
    \centering
    \caption{Node Classification Accuracy (\%) and Inference Time (ms) on Large-scale Graphs. ``$\uparrow$ $m~\times$'' indicates ProGMLP is $m$ times faster than the teacher at the inference stage.}
    \renewcommand\arraystretch{1.4}
    \begin{tabular}{l c c c c}
    \hline
    \toprule
          ~ & \multicolumn{2}{c}{ogbn-arxiv} & \multicolumn{2}{c}{ogbn-products} \\
          ~ & Acc.  & Inf. Time & Acc.  & Inf. Time \\
          \midrule
          GCN & 68.42$_{\pm0.67}$ & 84.13  & 72.94$_{\pm0.14}$ & 274.77\\
          \rowcolor[HTML]{E5F2FC}~+\textbf{ProGMLP} 
          & \textbf{70.11$_{\pm0.74}$}  & 3.95  & \textbf{73.00$_{\pm0.12}$} & 52.21\\
          
          \hdashline[3pt/3pt]
          \rowcolor[HTML]{FDECEE}\textbf{\emph{Impro.}} & +2.47\% & $\uparrow$ 21$\times$  & +0.08\% & $\uparrow$ 5$\times$ \\
          
          \midrule
          
          GAT & 69.25$_{\pm1.26}$ & 97.81 & OOM & - \\
          \rowcolor[HTML]{E5F2FC}~+\textbf{ProGMLP} 
          & \textbf{70.29$_{\pm1.94}$} & 12.80 & - & -\\
          
          \hdashline[3pt/3pt]
          \rowcolor[HTML]{FDECEE}\textbf{\emph{Impro.}} & +1.50\% & $\uparrow$ 8$\times$  & - & -\\
          
          \midrule

          GraphSAGE & 69.70$_{\pm0.21}$ & 27.99 & 75.59$_{\pm0.10}$ & 139.21\\
          \rowcolor[HTML]{E5F2FC}~+\textbf{ProGMLP} 
          & \textbf{70.97$_{\pm0.85}$} & 6.93 & \textbf{75.68$_{\pm0.07}$} & 59.06\\
          \hdashline[3pt/3pt]
          \rowcolor[HTML]{FDECEE}\textbf{\emph{Impro.}} & +1.80\% & $\uparrow$ 4$\times$  & +0.01\% & $\uparrow$ 2$\times$ \\
    \bottomrule
    \hline
    \end{tabular}
    \label{tab:ogb}
\end{table}

%
%
%
%
%
%
%
%

We evaluate the node classification performance of ProGMLP on two large-scale graph datasets: ogbn-arxiv and ogbn-products. The evaluation focuses on both accuracy and inference time, highlighting the efficiency of ProGMLP in large-scale scenarios.

On the ogbn-arxiv dataset, ProGMLP consistently outperforms all teacher models in terms of accuracy. It improves by 2.47\% over GCN, 1.50\% over GAT, and 1.80\% over GraphSAGE. Additionally, ProGMLP significantly reduces inference time, being 21× faster than GCN, 8× faster than GAT, and 4× faster than GraphSAGE. These improvements show that ProGMLP efficiently balances high accuracy with reduced inference time, making it highly suitable for large-scale graph tasks where computational costs are critical. On the ogbn-products dataset, ProGMLP continues to demonstrate efficiency. Although the accuracy gains are smaller, with 0.08\% improvement over GCN and 0.01\% over GraphSAGE, the model excels in inference speed. ProGMLP achieves inference times that are 5× faster than GCN and 2× faster than GraphSAGE, confirming its advantage in scenarios where fast inference is required without sacrificing performance.

In summary, ProGMLP consistently improves accuracy while drastically reducing inference time on large-scale datasets. This makes ProGMLP a strong candidate for large-scale node classification tasks where both performance and computational efficiency are essential.

\subsection{Evaluation on Inductive Setting}
\label{sec:ind}
\begin{table}[!th]
    \centering
    \caption{Inductive Node Classification Accuracy (\%).}
    \renewcommand\arraystretch{1.4}
    \begin{tabular}{l c c c c }
    \hline
    \toprule
          ~ &  Cora & Pubmed & CS & Physics \\
          \midrule
          GCN & 70.31$_{\pm0.54}$ & 80.06$_{\pm0.37}$ & 90.93$_{\pm0.35}$ & 93.27$_{\pm0.26}$ \\
          \rowcolor[HTML]{E5F2FC}~+\textbf{ProGMLP} 
          & \textbf{73.07$_{\pm0.66}$}  & \textbf{87.95$_{\pm0.36}$} & \textbf{93.38$_{\pm0.31}$} & \textbf{95.44$_{\pm0.18}$} \\
          
          \hdashline[3pt/3pt]
          \rowcolor[HTML]{FDECEE}\textbf{\emph{Impro.}} & +3.92\% & +9.85\% & +2.69\% & +2.32\%  \\
          
          \midrule
          
          GAT & 71.47$_{\pm1.35}$ & 82.67$_{\pm0.88}$ & 90.51$_{\pm0.30}$ & 93.41$_{\pm0.18}$ \\
          \rowcolor[HTML]{E5F2FC}~+\textbf{ProGMLP} 
          & \textbf{72.50$_{\pm0.91}$} & \textbf{88.01$_{\pm0.42}$} & \textbf{94.02$_{\pm0.21}$} & \textbf{95.99$_{\pm0.20}$} \\
          
          \hdashline[3pt/3pt]
          \rowcolor[HTML]{FDECEE}\textbf{\emph{Impro.}} & +1.44\% & +6.46\% & +3.88\% & +2.76\%  \\
          
          \midrule
          
          GraphSAGE & 69.71$_{\pm1.13}$ & 85.56$_{\pm0.54}$ & 90.50$_{\pm0.53}$ & 93.95$_{\pm0.49}$ \\
          \rowcolor[HTML]{E5F2FC}~+\textbf{ProGMLP} 
          & \textbf{71.91$_{\pm0.71}$} & \textbf{88.60$_{\pm0.30}$} & \textbf{94.80$_{\pm0.21}$} & \textbf{95.26$_{\pm0.15}$} \\
          
          \hdashline[3pt/3pt]
          \rowcolor[HTML]{FDECEE}\textbf{\emph{Impro.}} & +3.16\% & +3.55\% & +4.75\% & +1.39\%  \\
    
    \bottomrule
    \hline
    \end{tabular}
    \label{tab:ind}
\end{table}

To evaluate the effectiveness of ProGMLP in scenarios where the model needs to generalize to new, unseen data, we conduct the inductive setting experiment across 4 graphs by following the same setting as in~\cite{adagmlp}.

The results presented in Table~\ref{tab:ind} demonstrate that ProGMLP consistently outperforms each of the baseline GNN models across all datasets. For instance, with the GCN teacher on the Pubmed dataset, ProGMLP achieves an accuracy of 87.95\%, reflecting a notable improvement over the baseline GCN’s 80.06\%. These gains could be attributed to ProGMLP’s unique mechanisms, particularly the PKD and PMA. The PKD approach allows each student model to incrementally refine the knowledge gained from its predecessor, helping to capture increasingly complex patterns in the data, which is especially effective when generalizing to new, unseen nodes in the inductive setting. Meanwhile, PMA facilitates the model's ability to learn robust feature representations by creating progressively more challenging synthetic samples, which strengthens the model’s generalization capabilities on inductive tasks. 

\subsection{Ablation Study}
\label{sec:ablation_study}
\begin{figure}[!htbp]
	\includegraphics[width=1\columnwidth]{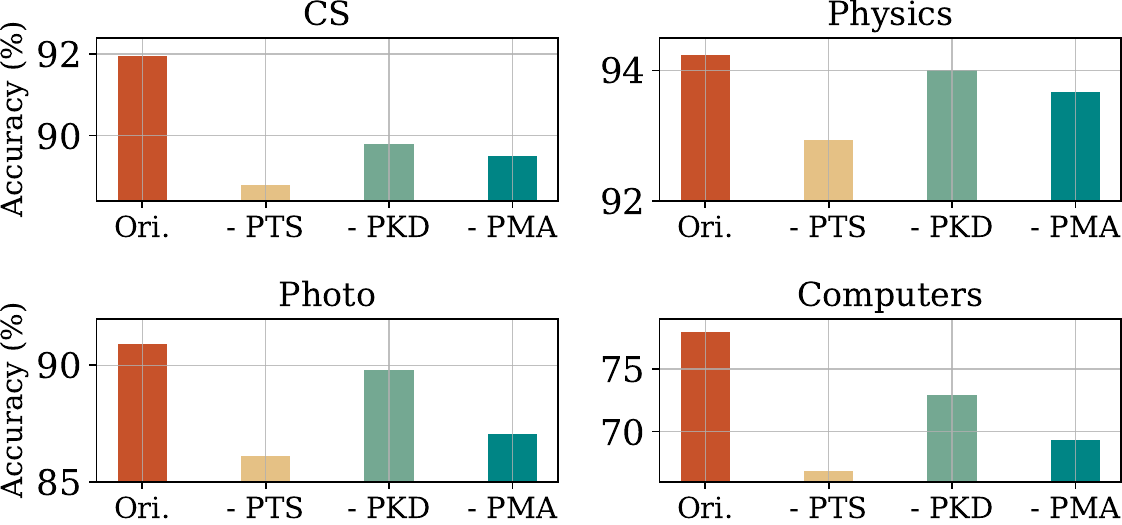}
	\caption{Ablation Study. Each figure compares the accuracy of the original ProGMLP (Ori.) with three ablated versions: without PTS (- PTS) / PKD (- PKD) / PMA (- PMA). \label{fig:abl}}
\end{figure}
To evaluate the impact of each major component in ProGMLP, we conduct an ablation study across four datasets. We progressively remove key components from the original ProGMLP model, including Progressive Training Structure (PTS), Progressive Knowledge Distillation (PKD), and Progressive Mixup Augmentation (PMA). By comparing the full model with ablated versions, we can identify the contributions of each component to the overall performance.

The results from Figure~\ref{fig:abl} show a significant performance drop when PTS is removed. For instance, in the CS dataset, accuracy decreases by approximately 4\% (from 92.43\% to 88.50\%) when PTS is excluded, indicating its crucial role in refining knowledge between successive MLP students. PTS is designed to allow each MLP student to progressively refine its knowledge from the previous student, which improves the overall model's ability to capture complex patterns. Without PTS, the later students are unable to build on previously learned knowledge, resulting in a significant decline in accuracy. When PKD is removed, there is a moderate decrease in accuracy, highlighting its role in effective knowledge transfer from the GNN teacher to the MLP students. The impact of removing PKD is less pronounced than PTS because PKD mostly aids in enhancing the learning efficiency, while PTS fundamentally changes the model's capacity to adapt to progressively harder tasks. Removing PMA also results in accuracy drops, particularly in the Computers dataset (from 77.92\% to 69.31\%). PMA introduces harder examples through mixup augmentation, which aids generalization by forcing the student models to learn more robust representations. While PMA is important for improving model generalization, its absence does not affect the core ability of the model to learn from progressively refined students, explaining why the drop is less severe compared to removing PTS.

\subsection{Hyperparameter Analysis}
\label{sec:hyper_analysis}
\begin{figure}[!htbp]
	\includegraphics[width=1\columnwidth]{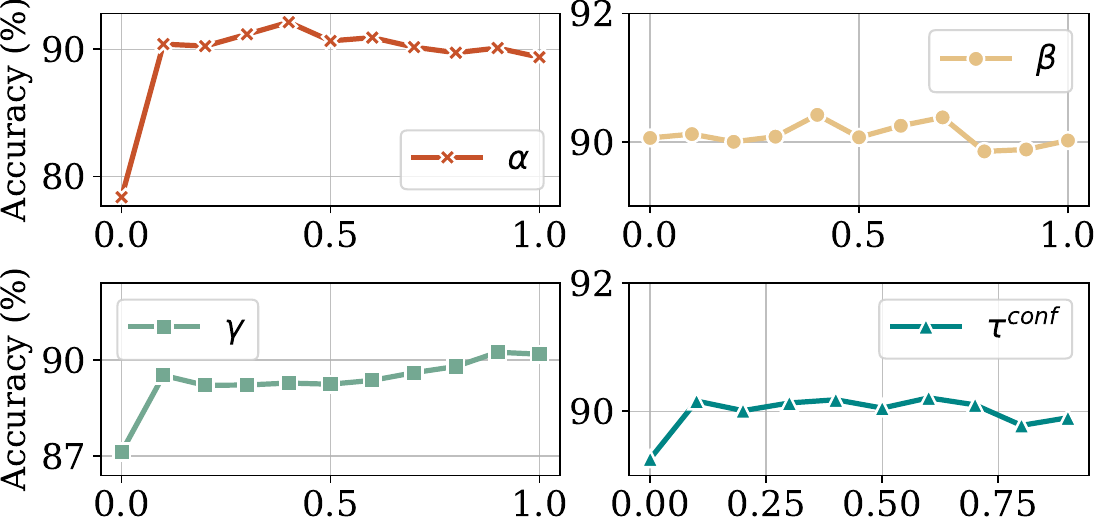}
	\caption{Hyperparameter Sensitivity Analysis.\label{fig:param}}

\end{figure}
We have conducted a hyperparameter study to analyze the sensitivity of ProGMLP to key hyperparameters on the CS dataset using the GCN as the teacher. The results are shown in Figure~\ref{fig:param}.

As we adjust $\alpha$, which balances the contributions of the Cross-Entropy loss and Knowledge Distillation loss, we can observe significant variations in accuracy. The model achieves its highest accuracy when $\alpha=0.4$, with performance peaking at 92.13\%. This shows that carefully balancing the supervision from true labels and the soft targets provided by the GNN teacher is critical for ProGMLP’s performance. Over-emphasizing either component leads to a drop in accuracy, demonstrating the importance of a balanced approach.

We also evaluate the influence of $\beta$, which increases the importance of the loss as we progress to later students in the sequence. The model’s accuracy remains relatively stable across different values of $\beta$, indicating that ProGMLP can tolerate variations in the strength of this effect without significant performance degradation. On the other hand, adjusting $\gamma$ produces only minor fluctuations in accuracy except for $\gamma=0$. Similarly, varying $\tau^{conf}$, the confidence threshold for early exit, shows that the model is robust across a range of confidence levels, maintaining efficient inference without significant accuracy loss.

\section{Conclusion}
In this paper, we present ProGMLP, a novel framework that bridges the gap between the high expressiveness of GNNs and the computational efficiency of MLPs. ProGMLP introduces a progressive learning mechanism that allows for flexible and adaptive trade-offs between inference cost and accuracy. Through comprehensive evaluations on multiple datasets, including large-scale graphs, ProGMLP demonstrates significant improvements in accuracy over existing methods while drastically reducing inference times. The results highlight ProGMLP’s suitability for real-world applications where computational resources and time are limited. The framework's ability to scale to large datasets further shows its effectiveness in balancing performance with efficiency. 

\section*{Acknowledgment}
This work was supported in part by the National Natural Science Foundation of China under Grants 62303366, 62425605, and 62133012, in part by the Key Research and Development Program of Shaanxi under Grants 2025CY-YBXM-041, 2022ZDLGY01-10, and 2024CY2-GJHX-15, and in part by the Fundamental Research Funds for the Central Universities under Grants ZYTS25086 and ZYTS25211.

\bibliographystyle{IEEEtran}
\bibliography{refs}

\begin{thebibliography}{10}
\providecommand{\url}[1]{#1}
\csname url@samestyle\endcsname
\providecommand{\newblock}{\relax}
\providecommand{\bibinfo}[2]{#2}
\providecommand{\BIBentrySTDinterwordspacing}{\spaceskip=0pt\relax}
\providecommand{\BIBentryALTinterwordstretchfactor}{4}
\providecommand{\BIBentryALTinterwordspacing}{\spaceskip=\fontdimen2\font plus
\BIBentryALTinterwordstretchfactor\fontdimen3\font minus \fontdimen4\font\relax}
\providecommand{\BIBforeignlanguage}[2]{{%
\expandafter\ifx\csname l@#1\endcsname\relax
\typeout{** WARNING: IEEEtran.bst: No hyphenation pattern has been}%
\typeout{** loaded for the language `#1'. Using the pattern for}%
\typeout{** the default language instead.}%
\else
\language=\csname l@#1\endcsname
\fi
#2}}
\providecommand{\BIBdecl}{\relax}
\BIBdecl

\bibitem{gcn}
T.~N. Kipf and M.~Welling, ``Semi-supervised classification with graph convolutional networks,'' \emph{arXiv preprint arXiv:1609.02907}, 2016.

\bibitem{gat}
P.~Veli{\v{c}}kovi{\'c}, G.~Cucurull, A.~Casanova, A.~Romero, P.~Lio, and Y.~Bengio, ``Graph attention networks,'' \emph{arXiv preprint arXiv:1710.10903}, 2017.

\bibitem{sage}
W.~Hamilton, Z.~Ying, and J.~Leskovec, ``Inductive representation learning on large graphs,'' in \emph{Advances in neural information processing systems}, 2017, pp. 1024--1034.

\bibitem{sgc}
F.~Wu, A.~Souza, T.~Zhang, C.~Fifty, T.~Yu, and K.~Weinberger, ``Simplifying graph convolutional networks,'' in \emph{International conference on machine learning}.\hskip 1em plus 0.5em minus 0.4em\relax PMLR, 2019, pp. 6861--6871.

\bibitem{gin}
K.~Xu, W.~Hu, J.~Leskovec, and S.~Jegelka, ``How powerful are graph neural networks?'' \emph{arXiv preprint arXiv:1810.00826}, 2018.

\bibitem{appnp}
J.~Klicpera, A.~Bojchevski, and S.~G{\"u}nnemann, ``Predict then propagate: Graph neural networks meet personalized pagerank,'' \emph{arXiv preprint arXiv:1810.05997}, 2018.

\bibitem{yang2022graph}
Y.~Yang, Z.~Guan, W.~Zhao, W.~Lu, and B.~Zong, ``Graph substructure assembling network with soft sequence and context attention,'' \emph{IEEE Transactions on Knowledge and Data Engineering}, vol.~35, no.~5, pp. 4894--4907, 2022.

\bibitem{skipnode}
W.~Lu, Y.~Zhan, B.~Lin, Z.~Guan, L.~Liu, B.~Yu, W.~Zhao, Y.~Yang, and D.~Tao, ``Skipnode: On alleviating performance degradation for deep graph convolutional networks,'' \emph{IEEE Transactions on Knowledge and Data Engineering}, pp. 1--14, 2024.

\bibitem{glnn}
S.~Zhang, Y.~Liu, Y.~Sun, and N.~Shah, ``Graph-less neural networks: Teaching old mlps new tricks via distillation,'' in \emph{International Conference on Learning Representations}, 2021.

\bibitem{nosmog}
Y.~Tian, C.~Zhang, Z.~Guo, X.~Zhang, and N.~Chawla, ``Learning mlps on graphs: A unified view of effectiveness, robustness, and efficiency,'' in \emph{The Eleventh International Conference on Learning Representations}, 2022.

\bibitem{krd}
L.~Wu, H.~Lin, Y.~Huang, and S.~Z. Li, ``Quantifying the knowledge in gnns for reliable distillation into mlps,'' \emph{arXiv preprint arXiv:2306.05628}, 2023.

\bibitem{ffg2m}
L.~Wu, H.~Lin, Y.~Huang, T.~Fan, and S.~Z. Li, ``Extracting low-/high-frequency knowledge from graph neural networks and injecting it into mlps: An effective gnn-to-mlp distillation framework,'' \emph{arXiv preprint arXiv:2305.10758}, 2023.

\bibitem{adagmlp}
W.~Lu, Z.~Guan, W.~Zhao, and Y.~Yang, ``Adagmlp: Adaboosting gnn-to-mlp knowledge distillation,'' in \emph{Proceedings of the 30th ACM SIGKDD Conference on Knowledge Discovery and Data Mining}, 2024, pp. 2060--2071.

\bibitem{edge_computing1}
K.~Cao, Y.~Liu, G.~Meng, and Q.~Sun, ``An overview on edge computing research,'' \emph{IEEE access}, vol.~8, pp. 85\,714--85\,728, 2020.

\bibitem{edge_computing2}
Y.~Mao, C.~You, J.~Zhang, K.~Huang, and K.~B. Letaief, ``A survey on mobile edge computing: The communication perspective,'' \emph{IEEE communications surveys \& tutorials}, vol.~19, no.~4, pp. 2322--2358, 2017.

\bibitem{edge_computing3}
J.~Chen and X.~Ran, ``Deep learning with edge computing: A review,'' \emph{Proceedings of the IEEE}, vol. 107, no.~8, pp. 1655--1674, 2019.

\bibitem{edge_computing4}
W.~Shi, J.~Cao, Q.~Zhang, Y.~Li, and L.~Xu, ``Edge computing: Vision and challenges,'' \emph{IEEE internet of things journal}, vol.~3, no.~5, pp. 637--646, 2016.

\bibitem{mobile1}
H.~Hoehle and V.~Venkatesh, ``Mobile application usability,'' \emph{MIS quarterly}, vol.~39, no.~2, pp. 435--472, 2015.

\bibitem{mobile2}
R.~Islam, R.~Islam, and T.~Mazumder, ``Mobile application and its global impact,'' \emph{International Journal of Engineering \& Technology}, vol.~10, no.~6, pp. 72--78, 2010.

\bibitem{early_exit1}
T.~Bolukbasi, J.~Wang, O.~Dekel, and V.~Saligrama, ``Adaptive neural networks for efficient inference,'' in \emph{International Conference on Machine Learning}.\hskip 1em plus 0.5em minus 0.4em\relax PMLR, 2017, pp. 527--536.

\bibitem{early_exit2}
D.~Dennis, C.~Pabbaraju, H.~V. Simhadri, and P.~Jain, ``Multiple instance learning for efficient sequential data classification on resource-constrained devices,'' \emph{Advances in Neural Information Processing Systems}, vol.~31, 2018.

\bibitem{anytime1}
A.~Ruiz and J.~Verbeek, ``Anytime inference with distilled hierarchical neural ensembles,'' in \emph{Proceedings of the AAAI Conference on Artificial Intelligence}, vol.~35, no.~11, 2021, pp. 9463--9471.

\bibitem{anytime2}
G.~Huang, D.~Chen, T.~Li, F.~Wu, L.~Van Der~Maaten, and K.~Q. Weinberger, ``Multi-scale dense networks for resource efficient image classification,'' \emph{arXiv preprint arXiv:1703.09844}, 2017.

\bibitem{anytime3}
D.~Dennis, A.~Shetty, A.~P. Sevekari, K.~Koishida, and V.~Smith, ``Progressive ensemble distillation: building ensembles for efficient inference,'' \emph{Advances in Neural Information Processing Systems}, vol.~36, pp. 43\,525--43\,543, 2023.

\bibitem{gcnii}
M.~Chen, Z.~Wei, Z.~Huang, B.~Ding, and Y.~Li, ``Simple and deep graph convolutional networks,'' in \emph{International conference on machine learning}.\hskip 1em plus 0.5em minus 0.4em\relax PMLR, 2020, pp. 1725--1735.

\bibitem{jknet}
K.~Xu, C.~Li, Y.~Tian, T.~Sonobe, K.-i. Kawarabayashi, and S.~Jegelka, ``Representation learning on graphs with jumping knowledge networks,'' in \emph{International conference on machine learning}.\hskip 1em plus 0.5em minus 0.4em\relax PMLR, 2018, pp. 5453--5462.

\bibitem{pcl}
W.~Lu, Z.~Guan, W.~Zhao, Y.~Yang, Y.~Lv, L.~Xing, B.~Yu, and D.~Tao, ``Pseudo contrastive learning for graph-based semi-supervised learning,'' \emph{arXiv preprint arXiv:2302.09532}, 2023.

\bibitem{nodemixup}
W.~Lu, Z.~Guan, W.~Zhao, Y.~Yang, and L.~Jin, ``Nodemixup: Tackling under-reaching for graph neural networks,'' in \emph{Proceedings of the AAAI Conference on Artificial Intelligence}, vol.~38, no.~13, 2024, pp. 14\,175--14\,183.

\bibitem{lpformer}
H.~Shomer, Y.~Ma, H.~Mao, J.~Li, B.~Wu, and J.~Tang, ``Lpformer: An adaptive graph transformer for link prediction,'' in \emph{Proceedings of the 30th ACM SIGKDD Conference on Knowledge Discovery and Data Mining}, 2024, pp. 2686--2698.

\bibitem{san}
Y.~Yang, Z.~Guan, W.~Zhao, W.~Lu, and B.~Zong, ``Graph substructure assembling network with soft sequence and context attention,'' \emph{IEEE Transactions on Knowledge and Data Engineering}, vol.~35, no.~5, pp. 4894--4907, 2022.

\bibitem{lassance2020deep}
C.~Lassance, M.~Bontonou, G.~B. Hacene, V.~Gripon, J.~Tang, and A.~Ortega, ``Deep geometric knowledge distillation with graphs,'' in \emph{ICASSP 2020-2020 IEEE International Conference on Acoustics, Speech and Signal Processing (ICASSP)}.\hskip 1em plus 0.5em minus 0.4em\relax IEEE, 2020, pp. 8484--8488.

\bibitem{zhang2023iterative}
H.~Zhang, S.~Lin, W.~Liu, P.~Zhou, J.~Tang, X.~Liang, and E.~P. Xing, ``Iterative graph self-distillation,'' \emph{IEEE Transactions on Knowledge and Data Engineering}, 2023.

\bibitem{ren2021multi}
Y.~Ren, J.~Ji, L.~Niu, and M.~Lei, ``Multi-task self-distillation for graph-based semi-supervised learning,'' \emph{arXiv preprint arXiv:2112.01174}, 2021.

\bibitem{joshi2022representation}
C.~K. Joshi, F.~Liu, X.~Xun, J.~Lin, and C.~S. Foo, ``On representation knowledge distillation for graph neural networks,'' \emph{IEEE Transactions on Neural Networks and Learning Systems}, 2022.

\bibitem{wu2022knowledge}
L.~Wu, H.~Lin, Y.~Huang, and S.~Z. Li, ``Knowledge distillation improves graph structure augmentation for graph neural networks,'' \emph{Advances in Neural Information Processing Systems}, vol.~35, pp. 11\,815--11\,827, 2022.

\bibitem{zhang2019your}
L.~Zhang, J.~Song, A.~Gao, J.~Chen, C.~Bao, and K.~Ma, ``Be your own teacher: Improve the performance of convolutional neural networks via self distillation,'' in \emph{Proceedings of the IEEE/CVF International Conference on Computer Vision}, 2019, pp. 3713--3722.

\bibitem{gnn-sd}
Y.~Chen, Y.~Bian, X.~Xiao, Y.~Rong, T.~Xu, and J.~Huang, ``On self-distilling graph neural network,'' \emph{arXiv preprint arXiv:2011.02255}, 2020.

\bibitem{kd_kl}
G.~Hinton, O.~Vinyals, and J.~Dean, ``Distilling the knowledge in a neural network,'' \emph{arXiv preprint arXiv:1503.02531}, 2015.

\bibitem{kd_l2}
J.~Ba and R.~Caruana, ``Do deep nets really need to be deep?'' \emph{Advances in neural information processing systems}, vol.~27, 2014.

\bibitem{lsp}
Y.~Yang, J.~Qiu, M.~Song, D.~Tao, and X.~Wang, ``Distilling knowledge from graph convolutional networks,'' in \emph{Proceedings of the IEEE/CVF Conference on Computer Vision and Pattern Recognition}, 2020, pp. 7074--7083.

\bibitem{tinygnn}
B.~Yan, C.~Wang, G.~Guo, and Y.~Lou, ``Tinygnn: Learning efficient graph neural networks,'' in \emph{Proceedings of the 26th ACM SIGKDD International Conference on Knowledge Discovery \& Data Mining}, 2020, pp. 1848--1856.

\bibitem{rdd}
\BIBentryALTinterwordspacing
W.~Zhang, X.~Miao, Y.~Shao, J.~Jiang, L.~Chen, O.~Ruas, and B.~Cui, ``Reliable data distillation on graph convolutional network,'' in \emph{Proceedings of the 2020 ACM SIGMOD International Conference on Management of Data}, ser. SIGMOD '20.\hskip 1em plus 0.5em minus 0.4em\relax New York, NY, USA: Association for Computing Machinery, 2020, p. 1399–1414. [Online]. Available: \url{https://doi.org/10.1145/3318464.3389706}
\BIBentrySTDinterwordspacing

\bibitem{gsdn}
L.~Wu, J.~Xia, H.~Lin, Z.~Gao, Z.~Liu, G.~Zhao, and S.~Z. Li, ``Teaching yourself: Graph self-distillation on neighborhood for node classification,'' \emph{arXiv preprint arXiv:2210.02097}, 2022.

\bibitem{vqgraph}
L.~Yang, Y.~Tian, M.~Xu, Z.~Liu, S.~Hong, W.~Qu, W.~Zhang, C.~Bin, M.~Zhang, and J.~Leskovec, ``Vqgraph: Rethinking graph representation space for bridging gnns and mlps,'' in \emph{The Twelfth International Conference on Learning Representations}, 2024.

\bibitem{early_exit3}
S.~Laskaridis, A.~Kouris, and N.~D. Lane, ``Adaptive inference through early-exit networks: Design, challenges and directions,'' in \emph{Proceedings of the 5th International Workshop on Embedded and Mobile Deep Learning}, 2021, pp. 1--6.

\bibitem{early_exit4}
S.~Teerapittayanon, B.~McDanel, and H.-T. Kung, ``Branchynet: Fast inference via early exiting from deep neural networks,'' in \emph{2016 23rd international conference on pattern recognition (ICPR)}.\hskip 1em plus 0.5em minus 0.4em\relax IEEE, 2016, pp. 2464--2469.

\bibitem{torch_geo}
M.~Fey and J.~E. Lenssen, ``Fast graph representation learning with {PyTorch Geometric},'' in \emph{ICLR Workshop on Representation Learning on Graphs and Manifolds}, 2019.

\bibitem{cora}
P.~Sen, G.~Namata, M.~Bilgic, L.~Getoor, B.~Galligher, and T.~Eliassi-Rad, ``Collective classification in network data,'' \emph{AI magazine}, vol.~29, no.~3, pp. 93--93, 2008.

\bibitem{coauthor}
O.~Shchur, M.~Mumme, A.~Bojchevski, and S.~G{\"u}nnemann, ``Pitfalls of graph neural network evaluation,'' \emph{arXiv preprint arXiv:1811.05868}, 2018.

\bibitem{ogb}
W.~Hu, M.~Fey, M.~Zitnik, Y.~Dong, H.~Ren, B.~Liu, M.~Catasta, and J.~Leskovec, ``Open graph benchmark: Datasets for machine learning on graphs,'' \emph{arXiv preprint arXiv:2005.00687}, 2020.

\bibitem{hgmd}
\BIBentryALTinterwordspacing
L.~Wu, Y.~Liu, H.~Lin, Y.~Huang, and S.~Z. Li, ``Teach harder, learn poorer: Rethinking hard sample distillation for gnn-to-mlp knowledge distillation,'' in \emph{Proceedings of the 33rd ACM International Conference on Information and Knowledge Management}, ser. CIKM '24.\hskip 1em plus 0.5em minus 0.4em\relax New York, NY, USA: Association for Computing Machinery, 2024, p. 2554–2563. [Online]. Available: \url{https://doi.org/10.1145/3627673.3679586}
\BIBentrySTDinterwordspacing

\end{thebibliography}

\end{document}